\documentclass{article}

\usepackage[final, nonatbib]{nips_2017}

\usepackage[utf8]{inputenc} 
\usepackage[T1]{fontenc}    
\usepackage{hyperref}       
\usepackage{url}            
\usepackage{booktabs}       
\usepackage{amsfonts}       
\usepackage{nicefrac}       
\usepackage{microtype}      
\usepackage[numbers]{natbib}
\usepackage{graphicx}
\graphicspath{ {images/} }

\makeatletter
\@addtoreset{footnote}{section}
\makeatother

\title{Improvised Comedy as a Turing Test}

\author{
  Kory Mathewson \thanks{Both authors contributed equally.} \\
  Department of Computing Science \\
  University of Alberta \\
  Edmonton, Alberta, Canada \\
  \texttt{korymath@gmail.com} \\
  \And
  Piotr Mirowski \mbox{\normalsize $^*$} \\
  HumanMachine \\
  London, UK \\
  \texttt{piotr.mirowski@computer.org} \\
}

\begin{document}

\maketitle

\begin{abstract}
The best improvisational theatre actors can make any scene partner, of any skill level or ability, appear talented and proficient in the art form, and thus "make them shine". To challenge this improvisational paradigm, we built an artificial intelligence (AI) trained to perform live shows alongside human actors for human audiences. Over the course of 30 performances to a combined audience of almost 3000 people, we have refined theatrical games which involve combinations of human and (at times, adversarial) AI actors. We have developed specific scene structures to include audience participants in interesting ways. Finally, we developed a complete show structure that submitted the audience to a Turing test and observed their suspension of disbelief, which we believe is key for human/non-human theatre co-creation.
\end{abstract}

\section{Background}
Theatrical improvisation is a form of live theatre where artists perform "real-time dynamic problem solving" through semi-structured spontaneous storytelling \cite{magerko2009empirical}. Improvised comedy involves both performers and audience members in interactive formats. 
We present explorations in a theatrical Turing Test as part of an improvised comedy show. We have developed an artificial intelligence-based improvisational theatre actor---a chatbot with speech recognition and speech synthesis, with a physical humanoid robot embodiment \cite{perlin1996improv, hayes1996improvisational} and performed alongside it in improv shows\footnote{Show listings and recordings are available at ~\url{https://humanmachine.live}} at performing arts festivals, including ImproFest UK and the Brighton, Camden, and Edinburgh Fringe Festivals \cite{mathewson2017improvised}.
Over these first 30 shows, one or two humans performed improvised scenes with the AI. The performers strove to endow the AI with human qualities of character/personality, relationship, status, emotion, perspective, and intelligence, according to common rules of improvisation \cite{johnstone2012impro, napier2004improvise}. Relying on custom state-of-the-art neural network software for language understanding and text generation, we were able to produce context-dependent replies for the AI actor. 

The system we developed aims to maintain the illusion of intelligent dialogue. Improvised scenes developed emotional connections between imaginary characters played by humans and AI improvisors. The human-like characterization elicited attachment for the AI from audience members. Through various configurations (e.g. human-human, human-AI, and AI-AI) and different AI embodiments (e.g. voice alone, visual avatar, or robot), we challenged the audience to discriminate between human- and AI-led improvisation. In one particular game setup, through a Wizard-of-Oz illusion, we performed a Turing test inspired structure. We deceived the audience into believing that an AI was performing, then we asked them to compare that performance with a performance by an actual AI. Feedback from the audience, and from performers who have experimented with our system, provide insight for future development of improv games. Below we present details on how we debuted this technology to audiences, and provide strictl anecdotal observations collected over multiple performances.


\section{Methods}
We named our AI improviser A.L.Ex, the Artificial Language Experiment, an homage to \textit{Alex the Parrot}, trained to communicate using a vocabulary of 150 words \cite{pepperberg2009alex}.
The core of A.L.Ex consists of a text-based chatbot implemented as a word-level sequence-to-sequence recurrent neural network (4-layer LSTM encoder, similar decoder, and topic model inputs) with an output vocabulary of 50k words. The network was trained on cleaned and filtered subtitles from about 100k films\footnote{Subtitles from 100k movies were collected from \url{https://opensubtitles.org}}.
Dialogue turn-taking, candidate sentence selection, and sentiment analysis \cite{hutto2014vader} on the input sentences are based on heuristics. The chatbot communicates with performers through out-of-the-box speech recognition and text-to-speech software. The chatbot runs on a local web server for modularity and allows for integration with physical embodiments (e.g. parallel control of a humanoid robot\footnote{The robot was manufactured by \url{https://www.ez-robot.com}}.
The server also enables remote connection which can override the chatbot and give dialog control to a human operator. Further technological implementation details are provided by \citet{mathewson2017improvised}.

An improvisational scene starts by soliciting suggestion for context from the audience (e.g., “non-geographical location” or “advice a grandparent might give”). The human performer then says several lines of dialogue to prime the AI with dense context. The scene continues through alternating lines of dialog between the human improviser(s) and the AI. Often through human justification, performers aim to maintain scene reality and ground narrative in believable storytelling. A typical scene lasts between 3-6 minutes, and is interrupted by the human performer when it reaches a natural ending (e.g. narrative conclusion or comical high point).


The first versions of the improvising artificial stage companions had their stage presence reduced to projected video and amplified sound. We evolved to physical embodiments (i.e. the humanoid robot) to project the attention of the performer(s) and audience on a material avatar. Our robotic performers are distinctly non-human in size, shape, material, actuation and lighting. We chose humanoid robotics because the more realistic an embodiment is the more comfortable humans often are with it; though comfort sharply drops when creatures have human-like qualities but are distinctly non-human \cite{mori1970uncanny}.



The performances at the Camden and Edinburgh Fringe festivals involved a Turing test inspired scene conducted with the willing audience. We performed the scene by first deceiving the audience into believing that an AI was performing (whereas the chatbot and the robot were controlled by a human); then we performed a second scene with an actual AI. In game (1), we explained the Turing test first, then performed the two scenes consecutively and finally asked the audience to discriminate, through a vote, which scene was AI-led. In a different game (2), we performed the Wizard-of-Oz scene and then immediately asked, in character and as part of the performance, if the audience suspected that a human was in control of the chatbot.

\section{Preliminary Observations and Conclusions}

We summarize here anecdotal observations from our performance. In game (1), nearly everyone identified the AI from the human. However, we noted that in game (2) approximately half the audience members believed that an AI was performing flawlessly alongside human improvisor(s). When not forewarned about the Turing test, the audience (of various ages and genders) was convinced that the dialog system understood the details of the scene and responded immediately and contextually. The propensity of this delusion is likely driven by several factors: the context within which they are viewing the deception, the lack of personal awareness of the current state-of-the-art AI abilities, and emotional connections with the scene. Post-show discussions with audience members confirmed that when a performer tells the audience that an AI is controlling the robot's dialogue, the audience members will trust this information. Being at an improvisational show, they expect to suspend disbelief and use their imagination. Most of them were also unaware of capabilities and limitations of state-of-the-art AI systems, which highlights the responsibility of the AI community to communicate progress in AI effectively and to effectively invite public understanding of AI ability. Finally, we observed that the introduction of a humanoid robot, with a human-like voice, increased the audiences' propensity to immerse themselves in the imaginative narrative presented to them.

We plan to conduct an experimental study of the audience beliefs in shared AI and human creativity\cite{schmidhuber2010formal}. We hope to better understand the way that audiences enjoy art when co-created by humans and AIs, to create better tools and mediums for human expression.



\bibliographystyle{unsrtnat}
\bibliography{references}

\begin{thebibliography}{10}
\providecommand{\natexlab}[1]{#1}
\providecommand{\url}[1]{\texttt{#1}}
\expandafter\ifx\csname urlstyle\endcsname\relax
  \providecommand{\doi}[1]{doi: #1}\else
  \providecommand{\doi}{doi: \begingroup \urlstyle{rm}\Url}\fi

\bibitem[Magerko et~al.(2009)Magerko, Manzoul, Riedl, Baumer, Fuller, Luther,
  and Pearce]{magerko2009empirical}
Brian Magerko, Waleed Manzoul, Mark Riedl, Allan Baumer, Daniel Fuller, Kurt
  Luther, and Celia Pearce.
\newblock An empirical study of cognition and theatrical improvisation.
\newblock In \emph{Proceedings of the seventh ACM conference on Creativity and
  cognition}, pages 117--126. ACM, 2009.

\bibitem[Perlin and Goldberg(1996)]{perlin1996improv}
Ken Perlin and Athomas Goldberg.
\newblock Improv: A system for scripting interactive actors in virtual worlds.
\newblock In \emph{Proceedings of the 23rd annual conference on Computer
  graphics and interactive techniques}, pages 205--216. ACM, 1996.

\bibitem[Hayes-Roth and Van~Gent(1996)]{hayes1996improvisational}
Barbara Hayes-Roth and Robert Van~Gent.
\newblock Improvisational puppets, actors, and avatars.
\newblock In \emph{Computer Game Developers Conference}, 1996.

\bibitem[Mathewson and Mirowski(2017)]{mathewson2017improvised}
Kory Mathewson and Piotr Mirowski.
\newblock Improvised theatre alongside artificial intelligences.
\newblock In \emph{AAAI Conference on Artificial Intelligence and Interactive
  Digital Entertainment}, 2017.

\bibitem[Johnstone(2012)]{johnstone2012impro}
Keith Johnstone.
\newblock \emph{Impro: Improvisation and the theatre}.
\newblock Routledge, 2012.

\bibitem[Napier(2004)]{napier2004improvise}
Mick Napier.
\newblock \emph{Improvise: Scene from the inside out}.
\newblock Heinemann Drama, 2004.

\bibitem[Pepperberg and Pepperberg(2009)]{pepperberg2009alex}
Irene~M Pepperberg and Irene~M Pepperberg.
\newblock \emph{The Alex studies: cognitive and communicative abilities of grey
  parrots}.
\newblock Harvard University Press, 2009.

\bibitem[Hutto and Gilbert(2014)]{hutto2014vader}
Clayton~J Hutto and Eric Gilbert.
\newblock Vader: A parsimonious rule-based model for sentiment analysis of
  social media text.
\newblock In \emph{Eighth international AAAI conference on weblogs and social
  media}, 2014.

\bibitem[Mori(1970)]{mori1970uncanny}
Masahiro Mori.
\newblock The uncanny valley.
\newblock \emph{Energy}, 7\penalty0 (4):\penalty0 33--35, 1970.

\bibitem[Schmidhuber(2010)]{schmidhuber2010formal}
J{\"u}rgen Schmidhuber.
\newblock Formal theory of creativity, fun, and intrinsic motivation
  (1990--2010).
\newblock \emph{IEEE Transactions on Autonomous Mental Development}, 2\penalty0
  (3):\penalty0 230--247, 2010.

\end{thebibliography}

\appendix{}

\section*{Supplementary Material}
\subsection*{Acknowledgments}
We would like to acknowledge the fantastic collaborators and supporters throughout the project, in particular Adam Meggido, Colin Mochrie, Matt Schuurman, Paul Blinov, Lana Cuthbertson, Patrick Pilarski, as well as Alessia Pannese, Stephen Davidson, Stuart Moses, Roisin Rae, John Agapiou, Arfie Mansfield, Luba Elliott, Shama Rahman, Holly Bartolo, Benoist Brucker, Charles Sabourdin, Katy Schutte and Steve Roe. Thank you to Rapid Fire Theatre for a space for new ideas to flourish.

\subsection*{Illustrations}

\begin{figure}[h]
\caption{System diagram of the Artificial Language Experiment (A.L.Ex).}
\centering
\includegraphics[width=0.7\textwidth]{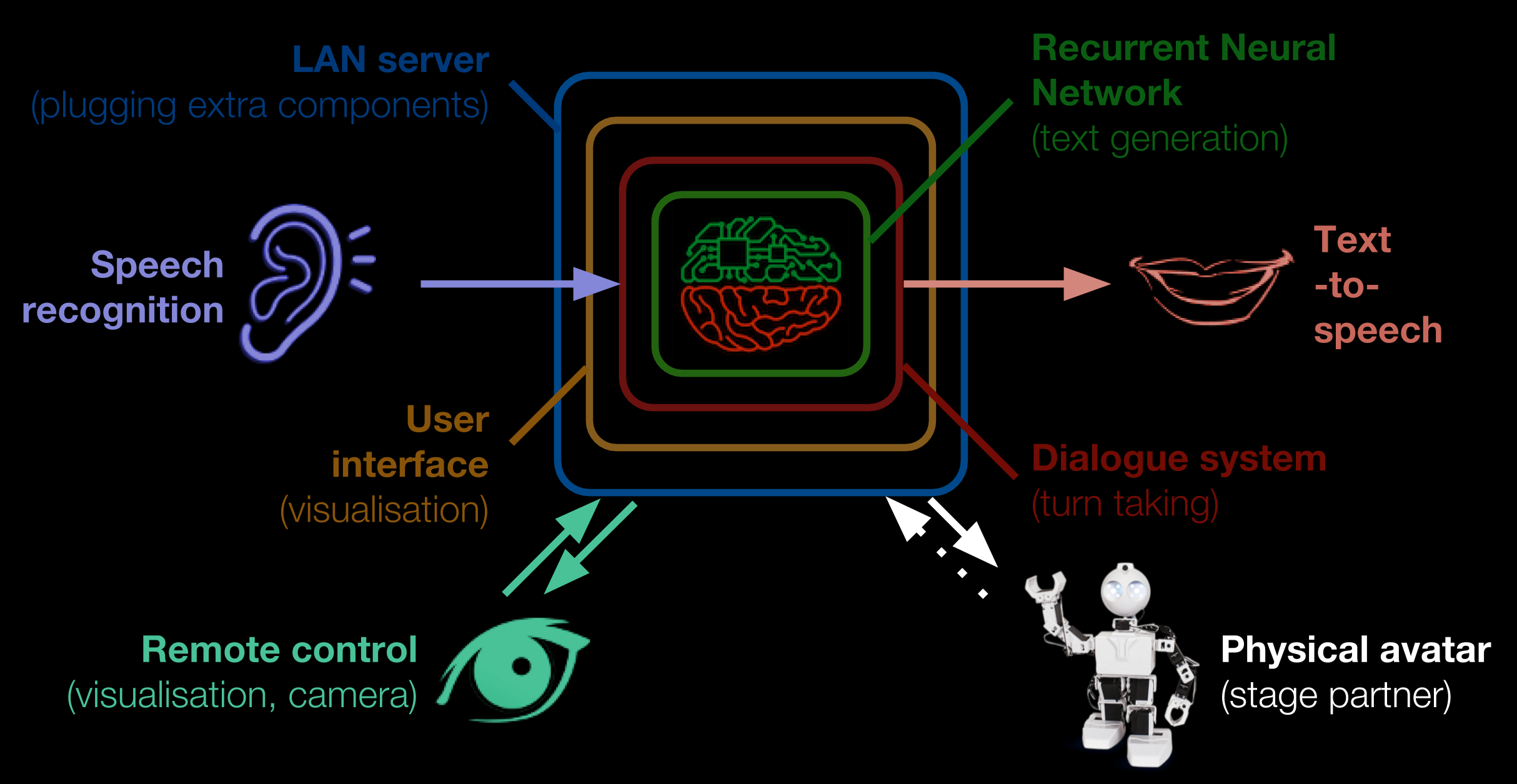}
\end{figure}

\begin{figure}[h]
\caption{Visual and physical embodiments of the AI improviser.}
\centering
\includegraphics[width=0.3\textwidth]{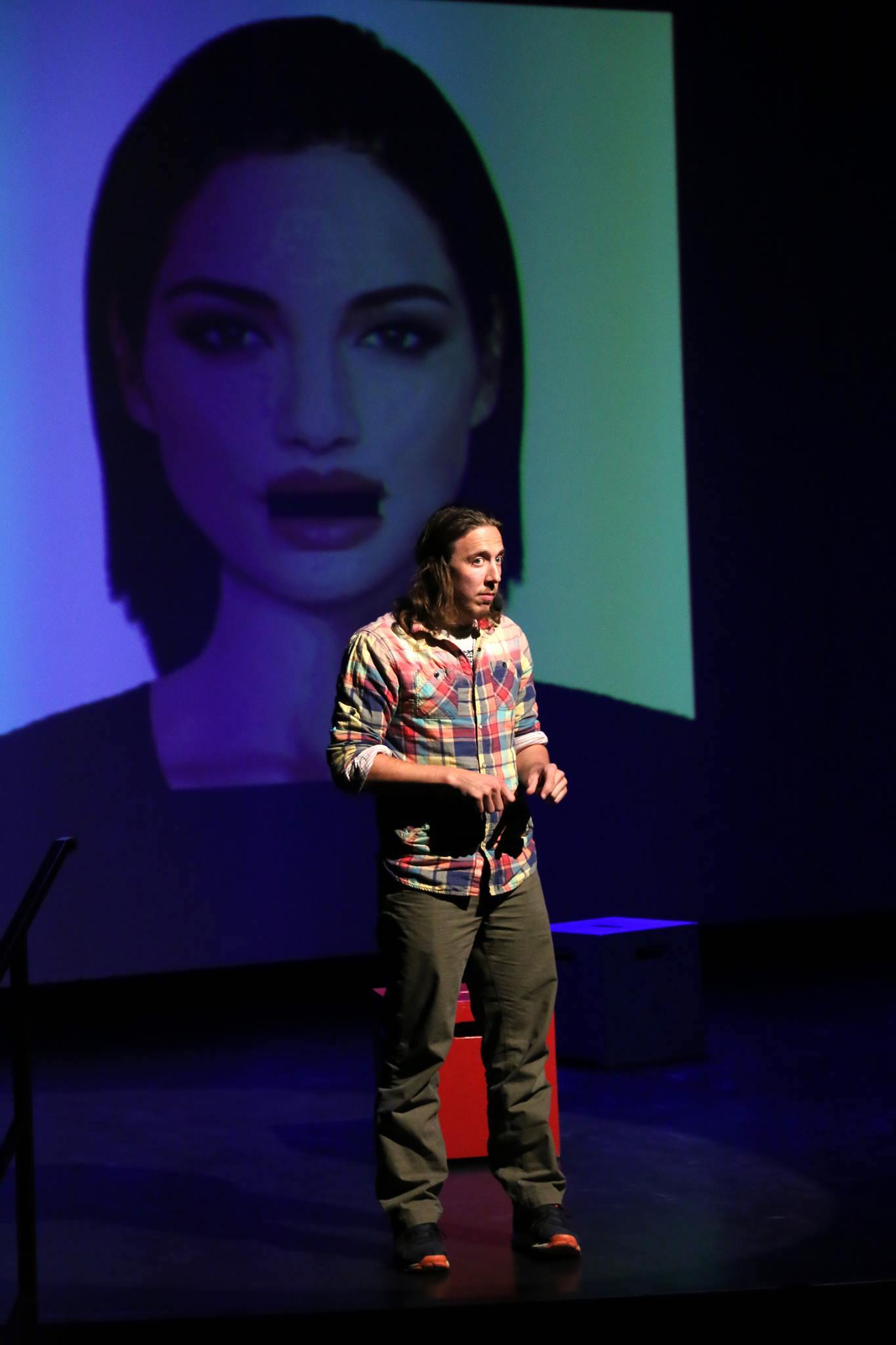}
\includegraphics[width=0.675\textwidth]{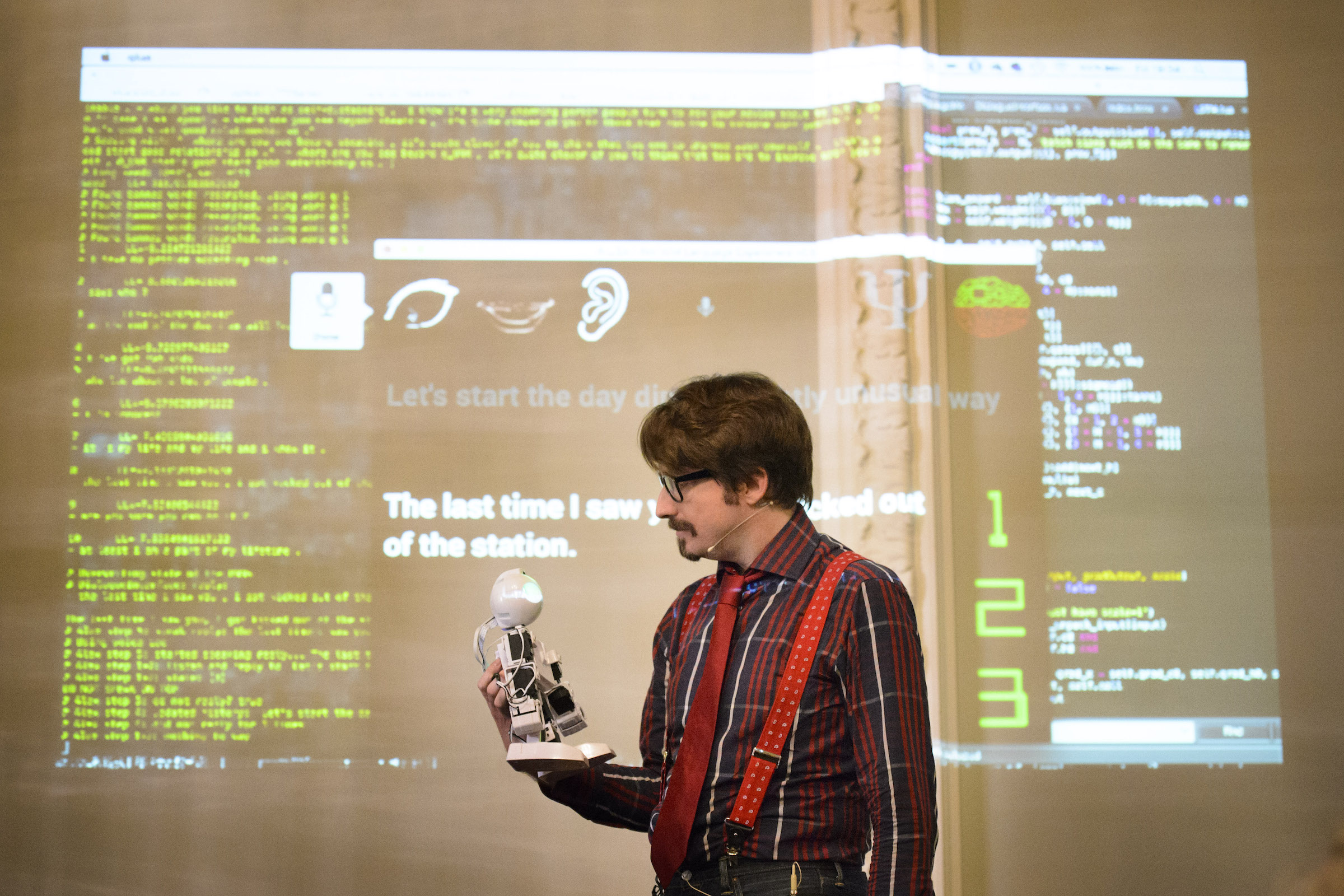}
\end{figure}

\begin{figure}[h]
\caption{Two human performers and an audience volunteer improvising with a robot.}
\centering
\includegraphics[width=0.8\textwidth]{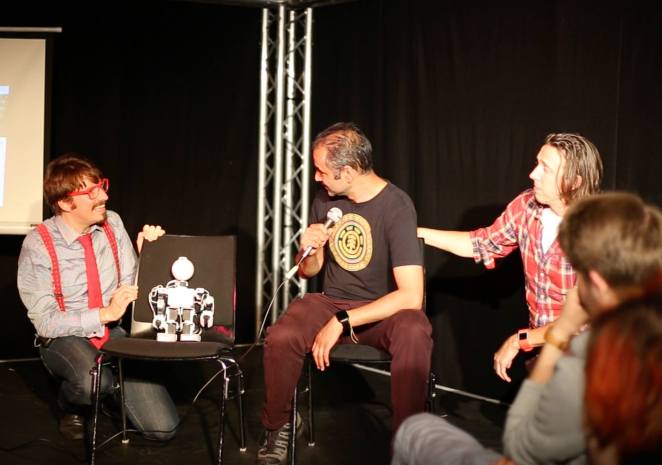}
\end{figure}

\end{document}